%% file: main.tex
\documentclass[runningheads]{llncs}
\usepackage[T1]{fontenc}
\usepackage{graphicx}
\usepackage{booktabs}
\usepackage{hyperref}
\usepackage{subfigure}
\usepackage{multirow}
\usepackage{amsmath}
\usepackage{amssymb}
\usepackage[capitalize]{cleveref}
\usepackage{colortbl}
\usepackage[inline,shortlabels]{enumitem}
\usepackage{etoolbox}
\usepackage{array}
\usepackage{booktabs}      
\usepackage{xcolor}        
\usepackage{pifont}        
\usepackage{soul}

\newcommand{\cmark}{\textcolor{green!70!black}{\ding{51}}}  
\newcommand{\xmark}{\textcolor{red}{\ding{55}}}             

\makeatletter
\newcommand\blfootnote[1]{%
  \begingroup
  \renewcommand\thefootnote{}%
  \footnotetext[0]{#1}%
  \endgroup
}
\makeatother

\definecolor{edf1}{rgb}{0, 0.102, 0.439}
\definecolor{edf2}{rgb}{0, 0.357, 0.733}
\definecolor{edf3}{rgb}{1, 0.627, 0.184}
\definecolor{edf4}{rgb}{0.996, 0.345, 0.082}
\definecolor{edf5}{RGB}{196, 214, 0}
\definecolor{edf6}{RGB}{80, 158, 47}

\colorlet{maincolor}{edf1}

\definecolor{bgcolor}{HTML}{ECF1FC}
\definecolor{fgcolor}{HTML}{222244}

\definecolor{colframe}{gray}{0.2}
\colorlet{colback}{bgcolor}

\usepackage[most]{tcolorbox}

\usepackage[misc]{ifsym}

\usepackage{mwe}

\begin{document}

\title{MoTM: Towards a Foundation Model for Time Series Imputation based on Continuous Modeling}
\titlerunning{MoTM: Towards a Foundation Model for TS Imputation}


\author{Etienne Le Naour\textsuperscript{*}, Tahar Nabil\textsuperscript{*}, Ghislain Agoua}

\authorrunning{E. Le Naour, T. Nabil, G. Agoua}

\institute{EDF R\&D \\ \email{\{name.surname\}@edf.fr}}

\maketitle              

\input{00-abstract}
\input{01-intro}
\input{02-related-works}
\input{03-methods}
\input{04-experiments}

\input{05-conclusion}

%
%
%
\bibliographystyle{splncs04}
\bibliography{main}

\end{document}

%% file: 00-abstract.tex
\begin{abstract}

Recent years have witnessed a growing interest for time series foundation models, with a strong emphasis on the forecasting task.
Yet, the crucial task of out-of-domain imputation of missing values remains largely underexplored.
We propose a first step to fill this gap by leveraging implicit neural representations (INRs).
INRs model time series as continuous functions and naturally handle various missing data scenarios and sampling rates.
While they have shown strong performance within specific distributions, they struggle under distribution shifts.
To address this, we introduce \textit{MoTM} (Mixture of Timeflow Models), a step toward a foundation model for time series imputation.
Building on the idea that a new time series is a mixture of previously seen patterns, \emph{MoTM} combines a basis of INRs, each trained independently on a distinct family of time series, with a ridge regressor that adapts to the observed context at inference.
We demonstrate robust in-domain and out-of-domain generalization across diverse imputation scenarios (e.g., block and pointwise missingness, variable sampling rates), paving the way for adaptable foundation imputation models. \url{https://github.com/EtienneLnr/MoTM}

\keywords{Imputation \and Foundation model \and Implicit Neural Representations.}
\end{abstract}

%% file: 01-intro.tex
\section{Introduction}

\blfootnote{* Equal contribution \\ 10th Workshop on Advanced Analytics and Learning on Temporal Data (AALTD), ECML 2025}

Real-world time series from domains such as healthcare, industry and climate science are often irregularly sampled or incomplete due to sensor failures and decentralized data collection \cite{schulz1997spectrum,clark2004population}. Reliable imputation is thus a critical first step toward downstream tasks like forecasting, classification, or anomaly detection.
Yet, while recent deep learning methods have advanced imputation performance \cite{cao2018brits,du2023saits}, they typically lack robustness to distribution shifts and fail to generalize to out-of-domain data.

Recently, zero-shot forecasting models have emerged in the time series community, enabling inference on unseen datasets without retraining. This shift has led to the rise of time series foundation models, offering key benefits: (i) a single, deployable model for diverse use cases, (ii) strong performance on new datasets, often exceeding supervised baselines, and (iii) emerging capabilities beyond simple memorization. While forecasting foundation models are well-studied \cite{das2024timesfm,woo2024moirai,ansari2024chronos}, imputation-focused counterparts remain scarce. Notable attempts like NuwaTS \cite{cheng2024nuwats} and MOMENT \cite{goswami2024moment} address imputation, but overlook data heterogeneity and varying sampling rates by relying on fixed-length input segments, limiting their ability to exploit shared patterns like periodicities across datasets.

A promising direction to overcome this issue lies in time series continuous-time modeling, recently advanced through the use of implicit neural representations (INR) \cite{DeepTime,naour2023time,li2025imputeinr}. These models enable time series to be represented as continuous functions, making them particularly suitable for imputation tasks involving irregular sampling or unaligned timestamps. Among these, TimeFlow \cite{naour2023time} has demonstrated competitive imputation performance, often matching or surpassing both traditional statistical methods and deep learning-based approaches. However, while TimeFlow excels within a specific data distribution, it struggles to generalize across distributions, limiting its utility in out-of-domain (OOD) settings.

To address these limitations, we introduce MoTM (Mixture of TimeFlow Models), a novel mixture-based architecture. MoTM leverages a ridge regression mechanism at inference to aggregate latent representations from multiple TimeFlow models, each trained on a distinct domain. Our contributions are as follows.

\begin{itemize}[label=\textbullet]
    \item We propose MoTM, a unified model capable of (i) handling various patterns of missing values at inference; (ii) achieving strong performance on out-of-domain datasets without retraining; (iii) effectively managing datasets sampled at different rates (e.g., 10min, 30min, 1h, 2h) by leveraging shared temporal structures across distributions. To the best of our knowledge, MoTM is the first model to meet all of these conditions (see \cref{tab:settings}).
    \item Our experiments on synthetic datasets reveal that MoTM exhibits strong zero-shot imputation capabilities that go beyond mere memorization. Notably, it generalizes effectively to time series subject to strong distribution shifts without any additional training.
    \item On real-world datasets, MoTM surpasses baseline models on in-domain (ID) inference and matches the performance of the strongest supervised approaches in the out-of-domain (OOD) setting.
\end{itemize}

\input{tables/table-settings}

%% file: tables/table-settings.tex
\begin{table}[h]
\centering
\caption{Comparison of imputation models on key generalization capabilities.}
\setlength{\tabcolsep}{6pt} 
\label{tab:settings}
\scalebox{0.80}{
\begin{tabular}{@{}lccc@{}}
\toprule
\multirow{2}{*}{\textbf{Method}} & \textbf{Can Impute Various} & \textbf{Natively Support Different} & \textbf{Can Perform } \\
  & \textbf{Missing Patterns} & \textbf{Sampling Rates} & \textbf{OOD Inference} \\
\midrule
BRITS \cite{cao2018brits},SAITS \cite{du2023saits} & \cmark & \xmark & \xmark \\
NuwaTS \cite{cheng2024nuwats},MOMENT \cite{goswami2024moment}  & \cmark & \xmark & \cmark \\
TimeFlow \cite{naour2023time}, ImputeINR \cite{li2025imputeinr} & \cmark & \cmark & \xmark \\
\textbf{MoTM (Ours)}             & \cmark & \cmark & \cmark \\
\bottomrule
\end{tabular}}
\end{table}

%% file: 02-related-works.tex
\newpage
\section{Related Work}

The task of imputing missing values in time series has been extensively studied, with approaches ranging from classical statistical methods to recent advances in deep learning. In this section, we review relevant lines of work, focusing on three major directions: supervised deep imputation models, continuous-time representations, and the recent emergence of time series foundation models. 

\paragraph{Supervised imputation.}
Recent advances in deep learning have led to an increasing number of models for time series imputation. BRITS \cite{cao2018brits} pioneered the use of bidirectional RNNs for imputation, while subsequent methods explored alternative architectures such as GANs \cite{luo2018multivariate,luo2019e2gan}, VAEs \cite{fortuin2020gp}, diffusion models \cite{tashiro2021csdi}, matrix factorization techniques \cite{TIDER}, and Transformer-based models like SAITS \cite{du2023saits}. Despite their success, these models assume regularly sampled data limiting their flexibility in real-world applications. Moreover, their generalization capabilities in out-of-domain (OOD) settings remain limited.

\paragraph{Continuous-time models.}
Continuous-time modeling has emerged as a promising approach to handle irregular sampling in time series. Gaussian Processes (GPs) \cite{RasmussenW06} naturally represent functions over continuous domains but often struggle with scalability and kernel selection \cite{GPpriorneeded}. Neural Processes (NPs) \cite{GarneloRMRSSTRE18,kim2019analysis} provide a more scalable alternative by parameterizing GPs through encoder-decoder architectures, yet they remain challenged by complex, high-frequency signals. More recent extensions use diffusion-based priors \cite{BilosRSNG23}, but these approaches can be sensitive to the number of input timestamps. Other directions involve latent ODEs \cite{BrouwerSAM19,corr/abs-1907-03907}, or attention-based methods like mTAN \cite{ShuklaM21}, which model irregular time series in a continuous domain. However, these models often fall short in imputation accuracy compared to their discrete-time counterparts. Implicit neural representations (INRs) \cite{SIREN} have recently gained traction as a more expressive and flexible framework for continuous modeling \cite{DeepTime,naour2023time,li2025imputeinr}, but existing models like TimeFlow \cite{naour2023time} still exhibit limited generalization capabilities across distributions.

\paragraph{Foundation models.}

The emergence of foundation models for time series represents a shift towards models capable of zero-shot generalization across diverse datasets. In the forecasting domain, recent models such as Chronos \cite{ansari2024chronos}, Moirai \cite{woo2024moirai}, and TimesFM \cite{das2024timesfm} have demonstrated strong performance without fine-tuning, enabling deployment in heterogeneous environments. In contrast, foundation models for imputation remain underexplored. NuwaTS \cite{cheng2024nuwats} and MOMENT \cite{goswami2024moment} are among the few attempts in this direction, but both rely on fixed-length segments and struggle with irregular sampling or variable-resolution datasets. These limitations restrict their ability to model shared temporal structures across datasets with diverse characteristics; highlighting the need for more flexible, distribution-aware imputation models.


%% file: 03-methods.tex
\section{The MoTM Framework}
\label{motm-framework}

\subsection{Problem Setting}

We formalize the generalizable imputation problem across heterogeneous time series datasets. Our goal is to learn a universal model capable of imputing missing values over time series that vary in sampling frequency, temporal alignment, and underlying distribution. We now describe the notations and the imputation objectives at both training and inference time.

\paragraph{Data notations.}  
We consider a collection of $N_{\text{train}}$ training datasets, denoted by $\mathcal{D}_{\text{train}} = \{\mathcal{D}_i\}_{i=1}^{N_{\text{train}}}$, where each dataset $\mathcal{D}_i$ consists of $n_i$ time series:  
$\mathcal{D}_i = \{(\mathbf{x}^{(i,j)}, \mathcal{T}_{\text{obs}}^{(i,j)})\}_{j=1}^{n_i}$.  

\begin{itemize}[label=\textbullet]
    \item $\mathbf{x}^{(i,j)} \in \mathbb{R}^{T_j}$ denotes the $j^{\text{th}}$ time series in dataset $i$, consisting of $T_j$ observed values.
    \item The temporal grid $\mathcal{T}_{\text{obs}}^{(i,j)} \subset [0, 1]$ is the set of $T_j$ observed timestamps associated with $\mathbf{x}^{(i,j)}$. To facilitate learning shared temporal patterns across datasets with varying sampling rates, we rescale all time grids to lie within the interval $[0,1]$. For example, if we consider a time period spanning four weeks, the first time point is mapped to 0 and the last to 1. This common temporal reference allows us to align heterogeneous time series, regardless of sampling frequency, missing values, or alignment issues.
\end{itemize}

A visual illustration of the notations is provided below in \cref{fig:notations-vizu}.

\input{plots/fig-notations-vizu}

\paragraph{Imputation Task.}  
\emph{During training}, the goal is to learn a unified model $f_{\theta}$ capable of predicting the value $x_t$ at any given time $t \in [0,1]$ for any time series from any training dataset in $\mathcal{D}_{\text{train}}$. \emph{At inference time}, we aim for two generalizations:
\begin{itemize}[label=\textbullet]
    \item \textit{In-Domain Generalization:} accurately impute values $x_t$ for new time series from the training datasets ($\mathcal{D} \subset \mathcal{D}_{\text{train}}$).
    \item \textit{Out-of-Domain (OOD) Generalization:} accurately impute values $x_t$ for entirely new datasets not seen during training ($\mathcal{D}_{\text{new}} \not\subset\mathcal{D}_{\text{train}}$).
\end{itemize}

\subsection{Key Components}

Our framework is articulated around three key components:
\begin{enumerate} 
    \item \textbf{Pretraining: learn a basis of TimeFlow models on the training corpus $D_{\text{train}}$.} Each dataset in the training collection is used to learn a distinct TimeFlow model. These models capture per-dataset specific temporal patterns and collectively form a representative basis of diverse dynamics.
    
    \item \textbf{At inference step 1: adapt the basis of TimeFlow models for the target time series.} For each trained model in the basis, we optimize a latent code to best fit the new target series. This results in a set of modulated Implicit Neural Representations, each proposing a reconstruction of the input series from its own perspective.
    
    \item \textbf{At inference step 2: fit the orchestrator, here a ridge regressor, on top of the basis of TimeFlow models.} We extract hidden representations from each modulated model and combine them to form a shared feature space. A ridge regression is then trained to linearly combine these features for final imputation.
\end{enumerate}

In the following sections, we will elaborate on each component of our method.

\paragraph{TimeFlow architecture.} 
TimeFlow \cite{naour2023time} is an Implicit Neural Representation (INR) model, meaning that it is a neural network capable of learning a parameterized continuous function of time
 \( f_{\theta} \colon t \in [0,1] \mapsto f_{\theta}(t) \in \mathbb{R} \) 
that approximates a discrete time series \( \mathbf{x}_t \) at any time \( t \in [0,1] \). 

In contrast to plain INRs, which are typically designed to represent a single function (e.g., one time series), TimeFlow is a \emph{generalizable} INR, able to model an entire collection \( (x^{(j)})_j \) of time series. This is achieved by incorporating per-sample modulations (through per-sample additive bias $\psi^{(j)}$) that condition the function \( f_{\theta} \) on each specific instance.
See \cref{fig:gene-INR} for a visualization.
We refer to the original TimeFlow paper for a detailed description of this mechanism \cite{naour2023time}.

\input{plots/fig-generalizable-inr}


\paragraph{Learning a basis of TimeFlow models.} For each training dataset \(\mathcal{D}_i \in \mathcal{D}_{\text{train}}\), we train a distinct TimeFlow model $f_{\theta^{(i)}}$. This model is an INR conditioned by modulations optimized through a latent code \(z^{(i,j)}\) specific to each time series \(j\) of dataset \(i\).
Formally, the ouput of a TimeFlow model for the \(j\)-th series in dataset \(i\) is defined as:
\begin{align}
\hat{x}_t^{(i,j)} = f_{\theta^{(i)}, h_{w^{(i)}}}(t; z^{(i,j)}) := f_{\theta^{(i)}, \psi^{(i,j)}}(t),
\end{align}
where \(\psi^{(i,j)} = h_{w^{(i)}}(z^{(i,j)})\) is the modulation vector conditioning the INR's biases and $h_{w}(.)$ is a hypernetwork.

Each distinct TimeFlow instance is trained to optimize the following objective:
\begin{align}
\min_{\theta^{(i)},\,w^{(i)},\,\{z^{(i,j)}\}} \sum_{j=1}^{n_i} \mathcal{L}_{\mathcal{T}_{\text{obs}}^{(i,j)}} \big( x_t^{(i,j)}, f_{\theta^{(i)}, h_{w^{(i)}}}(t; z^{(i,j)}) \big),
\end{align}
where \(\mathcal{L}_{\mathcal{T}_{obs}}\) is the reconstruction loss (e.g., MSE loss function) computed over observed time points \(t \in \mathcal{T}_{\text{obs}}\).
The result is a \textbf{basis of \(N_{\text{train}}\) TimeFlow models} \(\{f_{\theta^{(i)}, h_{w^{(i)}}}\}_{i=1}^{N_{\text{train}}}\), each capturing dynamics specific to dataset \(i\). 

\vspace{0.05cm}
\paragraph{Adapting the basis of TimeFlow models.}
At inference, the basis is adapted to a new dataset \(\mathcal{D}_{\text{new}} = \{(\mathbf{x}^{(j)}, \mathcal{T}_{\text{obs}}^{(j)})\}_{j=1}^{n_{\text{new}}}\) by optimizing, for each model \(i\), a latent code \(z^{(i,j)*}\) per series \(j\), keeping the shared parameters fixed:
\begin{align}
z^{(i,j)*} = \arg\min_{z} \mathcal{L}_{\mathcal{T}_{\text{obs}}^{(j)}} \big( x_t^{(j)}, f_{\theta^{(i)}, h_{w^{(i)}}}(t; z) \big).
\end{align}

Since TimeFlow is trained using a meta-learning approach \cite{CAVIA}, adapting it to new time series involves quickly computing the corresponding latent codes $z^{(i,j)*}$ for each model $i$ and new series $j$. This adaptation requires only a few optimization steps based on the observed context (see \cite{naour2023time}). This process can be viewed as an \textit{inner-loop} optimization, allowing the pretrained models to adjust efficiently to new samples.

This yields a family of modulated INRs \(\{f_{\theta^{(i)}, h_{w^{(i)}}}(t; z^{(i,j)*})\}_{i=1}^{N_{\text{train}}}\) providing different reconstructions of series \(j\).

\vspace{0.05cm}
\paragraph{Fitting the orchestrator.}

To combine predictions from the adapted basis, we extract a latent representation \(r^{(i,j)} \in \mathbb{R}^d\) from each modulated model \(i\) for each observed values of series \(j\), typically the last hidden layer of each modulated INR of the basis:
\begin{align}
r^{(i,j)}(t) := \text{hidden\_repr}\big(f_{\theta^{(i)}, h_{w^{(i)}}}( t ; z^{(i,j)*}) \big)  \quad \text{for} \; t \in \mathcal{T}_{obs}^{j}.
\end{align}

The representations obtained by each one of the $N_{\text{train}}$ TimeFlow instances for the different observed timesteps are concatenated:
\begin{align}
\mathbf{R}_{obs}^{(j)} = 
\begin{bmatrix}
r^{(1,j)}(t_1) & r^{(2,j)}(t_1) & \ldots & r^{(N_{\text{train}},j)}(t_1) & 1\\
r^{(1,j)}(t_2) & r^{(2,j)}(t_2) & \ldots & r^{(N_{\text{train}},j)}(t_2) & 1\\
\vdots         & \vdots         & \ddots & \vdots & \vdots\\
r^{(1,j)}(T_j) & r^{(2,j)}(T_j) & \ldots & r^{(N_{\text{train}},j)}(T_j) & 1
\end{bmatrix}
\in \mathbb{R}^{T_j \times (N_{\text{train}}\cdot d + 1)}.
\end{align}

Then, for each time series \( j \), a ridge regression model is independently fitted to predict the observed values of \( \mathbf{x}^{(j)} \in \mathbb{R}^{T_j \times 1}\) as a linear combination of the representations in \( \mathbf{R}_{obs}^{(j)} \):
\begin{align}
\hat{\mathbf{x}}^{(j)} = \mathbf{R}_{obs}^{(j)} \mathbf{W}^{(j)},
\end{align}
where \( \mathbf{W}^{(j)} \in \mathbb{R}^{(N_{\text{train}}\cdot d + 1) \times 1} \) contains the regression coefficients and the intercept.
A visualization of this process is shown in \cref{fig:orchestrator-vizu}. 

\input{plots/fig-orchestrator}

These parameters are obtained by solving the following regularized least-squares problem: 
\begin{align}
\mathbf{W}^{*(j)} = \underset{\mathbf{W}^{(j)}}{\text{arg min}} \| \mathbf{x}^{(j)} - \mathbf{R}_{obs}^{(j)}\mathbf{W}^{(j)} \|_2^2 + \lambda \| \mathbf{W}^{(j)} \|_2^2,
\end{align}

with $\lambda\geq0$.
This optimization admits a closed-form solution, making the computation both efficient and scalable, even when the number of inputs is large. As such, the regression step acts as an \emph{orchestrator}, learning how to best combine the outputs of multiple TimeFlow models to produce robust and generalizable imputations for unseen time series.

\vspace{0.5cm}

\textbf{At inference:} to predict for new target timestamps coordinates, we just have to build its representation $\mathbf{R}_{target}^{(j)}$ and compute $\mathbf{R}_{target}^{(j)} \mathbf{W}^{*(j)}$.


%% file: plots/fig-notations-vizu.tex
\begin{figure}
    \centering
    \includegraphics[width=0.99\linewidth]{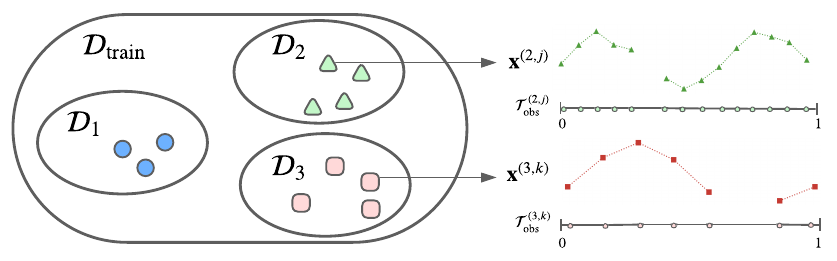}
    \caption{Illustration of the notations used in the rest of this paper.}
    \label{fig:notations-vizu}
\end{figure}

%% file: plots/fig-generalizable-inr.tex
\begin{figure}
    \centering
    \includegraphics[width=0.99\linewidth]{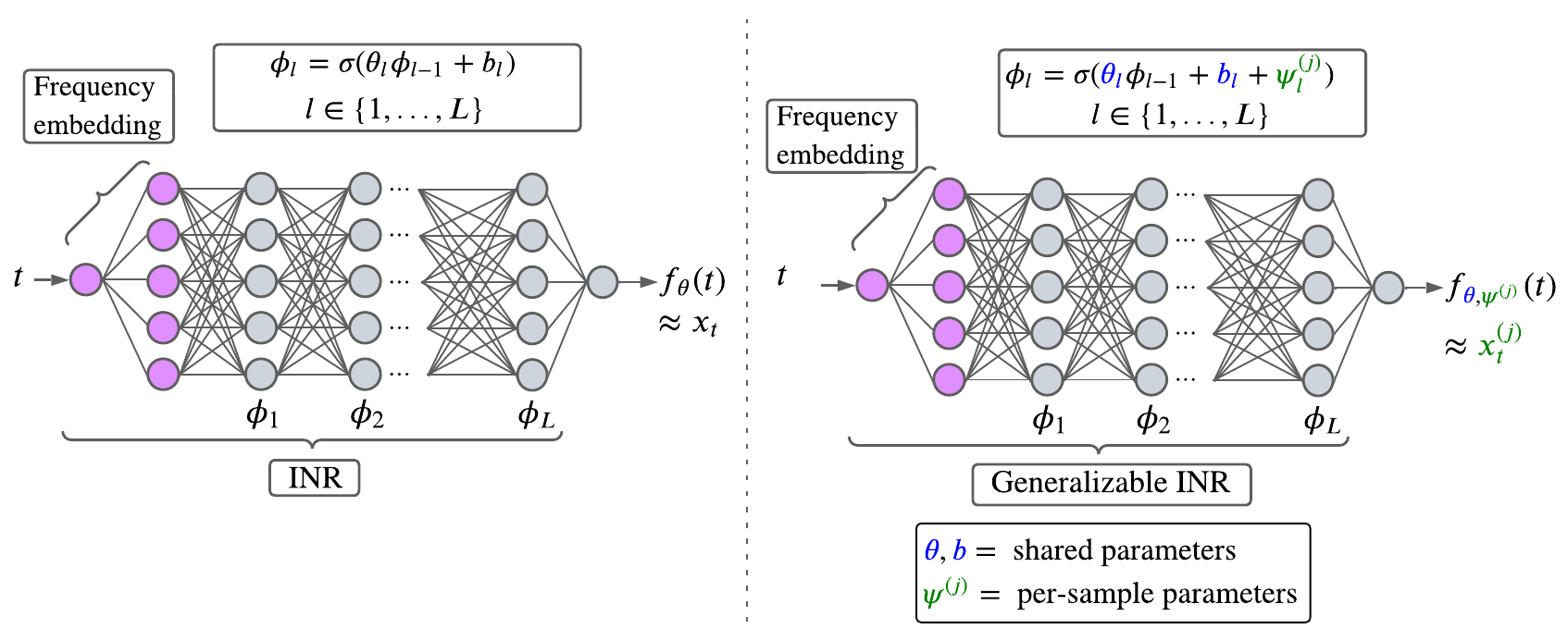}
    \caption{Plain INR vs Generalizable INR.}
    \label{fig:gene-INR}
\end{figure}

%% file: plots/fig-orchestrator.tex
\begin{figure}[h!]
    \centering
    \includegraphics[width=0.85\linewidth]{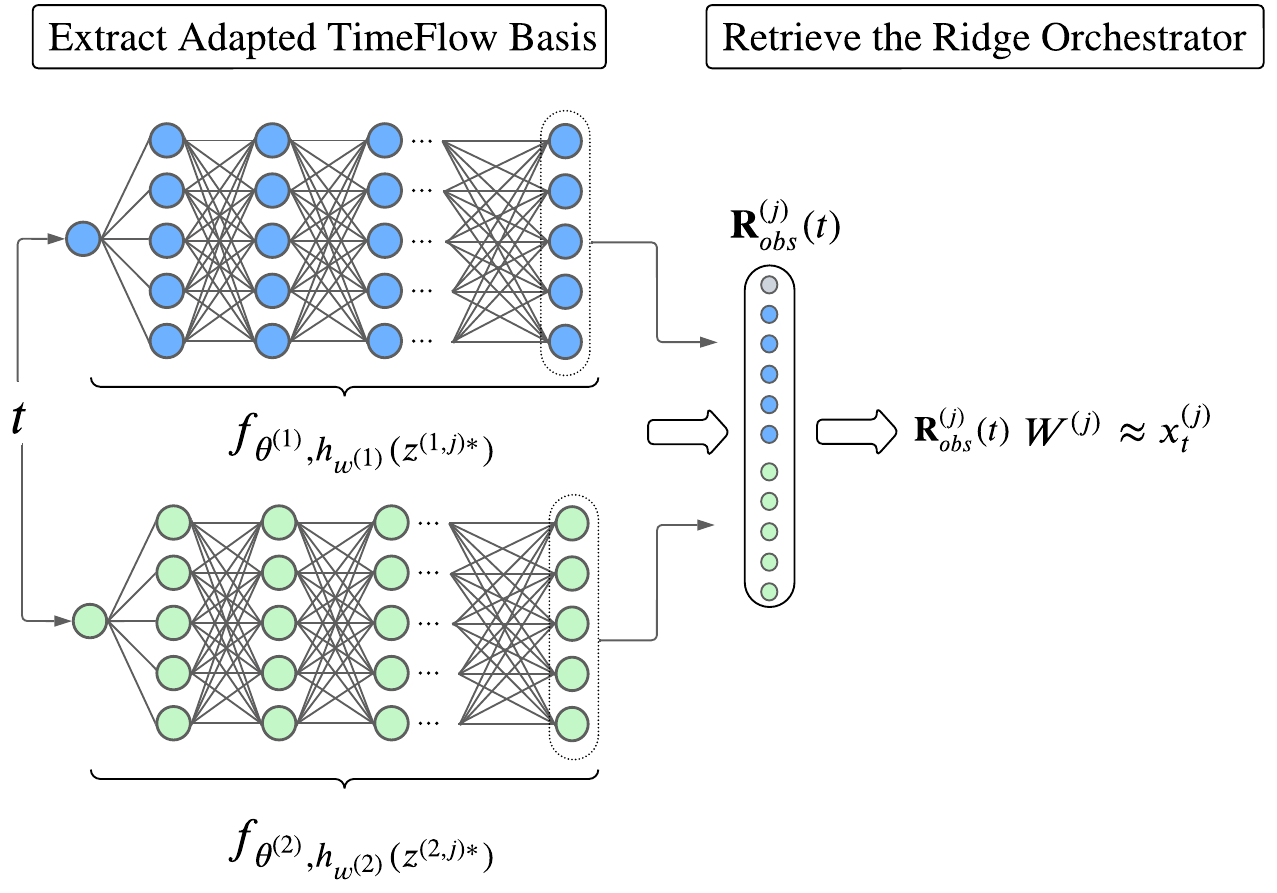}
    \caption{Illustration of how the ridge orchestrator operates on a new time series $\mathbf{x^{(j)}}$, using a basis of two TimeFlow models. Note that the linear projection matrix $W^{(j)}$ is jointly optimized over all observed time steps of the new series $\mathbf{x^{(j)}}$.}
    \label{fig:orchestrator-vizu}
\end{figure}

%% file: 04-experiments.tex
\section{Experiments}

We design two types of experiments to evaluate the performance of our proposed method. \begin{enumerate*}[(i)]
\item First, we consider controlled synthetic datasets that allow us to analyze the behavior of MoTM in a well-understood setting. 
\item Second, we evaluate MoTM on real-world datasets to assess the applicability and generalization capacity of our approach in more complex and diverse scenarios. 
\end{enumerate*} 

\subsection{Experiments on Synthetic Data} \label{ssec:synthetic}

We begin the empirical evaluation of MoTM by designing a synthetic experiment.
By controlling the seasonalities of the generated datasets, we aim to assess whether MoTM can generalize to an unseen combination of known patterns.

\paragraph{Data generating process.}
We generate three synthetic datasets summarized in \cref{tab:kernel-synth}, using  Gaussian Processes (GP) \emph{via} KernelSynth \cite{ansari2024chronos}.
Each synthetic series is the sum of three components, namely:
\begin{enumerate*}[label=(\roman*)]
\item a smooth trend sampled from a GP with RBF kernel;
\item a seasonal component sampled from a GP with a periodic Exponential-Sine-Squared kernel;
\item a residual sampled from a GP with white noise kernel.
\end{enumerate*}
We create $N_{\texttt{train}}=2$ datasets for pretraining, \texttt{ks1D} (resp., \texttt{ks1W}) with an hourly sampling frequency (resp., half-hourly) and daily (resp., weekly) periodicities.
A chronological 0.75 - 0.25 train - test split is applied on both \texttt{ks1D} and \texttt{ks1W}.
The third dataset \texttt{ks1D1W}, sampled at a 15-min rate with both a daily and a weekly periodic component, is kept for inference only.

\input{tables/table_kernel_synth}


\paragraph{Train protocol.}
Following the procedure described in \cref{motm-framework}, we train one TimeFlow model on \texttt{ks1D} and another on \texttt{ks1W}. Both models are trained on time grids spanning four-week periods. To condition each TimeFlow on partially observed time series, we simulate missing data during training by randomly removing a subset of timesteps from the grids $\mathcal{T}_{obs}$ at each optimization step, resulting in sparse inputs with a missingness ratio $\tau \in \{0.01,0.2,0.3,\dots,0.7\}$.
At inference time, as detailed in \cref{motm-framework}, MoTM reuses both pretrained TimeFlow models to fit a ridge regressor on new partially observed time series.

\paragraph{Test protocol.}
The test split is divided into four-week segments.
For each segment, we generate four distinct missing data scenarios, by randomly removing:
either
\begin{enumerate*}[label=(\roman*)]
\item 50\% (\emph{Point 1}) and
\item 70\% (\emph{Point 2}) of the observations;
or
\item two entire days (\emph{Block 1}) and
\item four entire days (\emph{Block 2}).
\end{enumerate*}


\paragraph{Implementation.}
TimeFlow's hyperparameters (see \cite{naour2023time}) are chosen as follows:
latent code of dimension 128, linear hypernetwork of size 256,
INR with a $2\times64$ frequency embedding of time, five 128-dimensional hidden layers.
The latent code is computed with 3 inner loop steps and a learning rate of 0.05.
We train with a batch size of 64 for $5\times10^3$ epochs, with a $10^{-3}$ learning rate for the INRs and hypernetworks.
After training, we set $\lambda=2$ for MoTM in all inference settings.

\paragraph{Baselines.}
We compare MoTM against several baselines:
$\bullet$ \texttt{TimeFlow 1D} and \texttt{TimeFlow 1W} respectively use predictions from a single TimeFlow model trained on \texttt{ks1D} and \texttt{ks1W}
$\bullet$ \texttt{Linear} performs standard linear interpolation between observed points
$\bullet$ \texttt{Repeat} imputes each missing value by copying the most recent available observation from the desired seasonnality.
Specifically, it uses the last available value from the previous day in the case of daily seasonality, or from the previous week in the case of weekly seasonality.
$\bullet$ \texttt{Mixture I} aggregates the predictions of both pretrained TimeFlow models using a softmax-weighted average, with weights derived from the negative reconstruction scores on the available context $\mathcal{T}^{j}_{obs}$
$\bullet$ \texttt{Mixture II} fits a ridge regressor on the observed context $\mathcal{T}^{j}_{obs}$, using the output predictions of both pretrained TimeFlow models as features.

\paragraph{Results.}
\cref{tab:res-synthetic} shows that MoTM achieves good performances on the in-domain datasets \texttt{ks1D} and \texttt{ks1W}, on a par with the strong supervised baselines.
We also note that \texttt{Mixture II}, the ridge regression on top of the TimeFlows predictions, emerges as a simple method to match the best pretrained model at inference.

\input{tables/table_synthetic_res}

Most notably, MoTM achieves strong results on \texttt{ks1D1W} in the OOD setting, highlighting its ability to generalize beyond simple memorization. Thanks to its continuous-time formulation, MoTM naturally adapts to the new sampling rate of \texttt{ks1D1W}, while the ridge orchestrator further improves performance, reducing the MAE by approximately 75\% compared to both pretrained TimeFlow variants. Indeed, the TimeFlow models trained individually on \texttt{ks1D} and \texttt{ks1W} transfer poorly to \texttt{ks1D1W}, confirming that TimeFlow can fit well within its training domain but struggles to generalize across distributions. In contrast, MoTM leverages the daily periodicity learned from \texttt{ks1D} and the weekly periodicity from \texttt{ks1W}, enabling it to handle the mixed \texttt{ks1D1W} dataset in a zero-shot manner.

Visually, \cref{fig:res-synthetic} showcases MoTM's ability to fit the context and impute missing values in challenging scenarios.

\input{figures/fig-synth-plots}

\subsection{Experiments on Real World Data}

In this section, we evaluate the performance of MoTM on various real-world datasets, covering both in-domain and out-of-domain scenarios.
Our goal is to assess its ability to generalize from a small set of heterogeneous datasets to a new target dataset.
Our experiments include comparisons with zero-shot, deep supervised, and statistical baselines under challenging imputation settings.





\paragraph{Datasets.}
For these experiments, the MoTM model is built on $N_{\text{train}} = 3$ TimeFlow models, each trained on one of the following datasets:
\begin{enumerate*}[label=(\roman*)]
\item \texttt{Electricity},
\item \texttt{Solar},
\item and \texttt{SpanishW-T}.
\end{enumerate*}
These datasets were selected to form a rich training domain, with diversity in sampling frequencies, seasonal patterns (daily and weekly for \texttt{Electricity}, daily for \texttt{Solar} and \texttt{SpanishW-T}), sample sizes (ranging from 5 to 321), and within-series variability.
For out-of-domain evaluation, we use standard benchmarks: \texttt{Traffic}, \texttt{ETTh1}, \texttt{ETTh2}, \texttt{Weather}, and \texttt{Spanish Energy}.
Their key characteristics are summarized in \cref{tab:data}.

\input{tables/table_data}

\paragraph{Protocol.}
All datasets are split chronologically intro train, validation and test fractions with respective ratios 0.7 - 0.1 - 0.2.
The ridge regularization coefficient $\lambda$ is selected from a grid search over $\{0.01,\,0.1,\,0.5,\,1,\,5,\,10\}$ on the validation split of the three training datasets.
The rest of the protocol for training and evaluation follows the one described for synthetic data.

\paragraph{Implementation.}
The hyperparameters for TimeFlow are the same as in \cref{ssec:synthetic}, except for an INR hidden size of 256.
We allow training for 40k epochs on \texttt{Electricity} and \texttt{Solar}, 20k epochs on \texttt{SpanishW-T}.
For MoTM, the grid search yields $\lambda=0.5$ for pointwise imputation and $\lambda=1$ for block imputation.

\paragraph{Baselines.}
\texttt{MoTM} is compared against three groups of baselines.
$\bullet$ We use \texttt{MOMENT} \cite{goswami2024moment} as another \emph{Zero-shot} foundation model, based on its Large version in inference mode across all datasets.
$\bullet$ The \emph{Supervised} baselines \texttt{TimeFlow} \cite{naour2023time}, \texttt{BRITS} \cite{cao2018brits} and \texttt{SAITS} \cite{du2023saits} are state-of-the-art deep imputation models trained on the respective train splits of each target dataset - including those tagged as OOD for MoTM.
$\bullet$ The \emph{Statistical} baselines are the \texttt{Linear} and \texttt{Repeat} interpolations described in the previous section.


\input{tables/table_v3}


\paragraph{Results.}
Several observations emerge from the test results reported in \cref{tab:res-clean}.
\begin{enumerate*}[label=(\roman*)]
\item MoTM consistently outperforms the zero-shot baseline MOMENT, which performs poorly across all datasets, highlighting the limitations of foundation models without adaptation.
\item MoTM achieves substantial gains over the supervised TimeFlow model on both in-domain (ID) and out-of-domain (OOD) datasets, demonstrating the benefits of multi-source training and the effectiveness of ridge-based adaptation.
\item Compared to other supervised deep learning methods, MoTM delivers competitive or superior performance — particularly against BRITS, reducing its error by $30.6\%$ on ID and $24.0\%$ on OOD datasets.
In addition, MoTM slightly outperforms SAITS in the ID setting ($12.6\%$ improvement on average), while remaining competitive in the OOD setting, even surpassing SAITS on datasets like \texttt{ETTh1} or \texttt{Weather} in the \emph{Block} scenarios.
\item MoTM also clearly outperforms statistical baselines such as linear interpolation and value repetition, with an average relative improvement of $22.5\%$ and $32.2\%$ in-domain, and $13.2\%$ and $31.1\%$ out-of-distribution.
\end{enumerate*}

Overall, the results highlight the benefits of continuous modeling for time series imputation, as well as the effectiveness of MoTM's simple adaptation strategy to generalize to new domains. Qualitatively,  \cref{fig:spanishE-vizu} illustrates a block imputation example from the OOD dataset \texttt{SpanishE}. Visually, MoTM demonstrates strong imputation capabilities on this complex sample.

\input{plots/fig-energy}

\paragraph{Ablations on the number of TimeFlow components in the MoTM basis.}
Can the performance of MoTM be matched by a single TimeFlow with ridge adaptation at inference?
To elucidate this question, we perform an ablation on $N_{\texttt{train}}$.
\cref{fig:ablation} shows the evolution of the test MAE on four datasets as the mixture grows from one component with ridge adaptation (\texttt{Electricity}) to all three components.
In general, increasing $N_{\texttt{train}}$ does improve the test metrics on ID and OOD datasets, showing that MoTM leverages its multi-source pretraining.
However, certain datasets and settings such as block imputation on \texttt{Traffic}, do not benefit from more base components.
This calls for further improvement of the orchestration mechanism, e.g. through a better tuning of $\lambda$.

\input{figures/fig-ablation_ntrain}

\paragraph{Discussion on inference time.} 
At first glance, the inference procedure of MoTM may appear computationally expensive, due to the combination of adapting TimeFlow models and fitting the ridge regression. 
To assess its practical cost, we report in \cref{tab:infer-time} the inference time of MoTM on the largest test dataset, \texttt{Traffic}. On a single H100 GPU, imputing 83k segments of length 672 takes approximately 61 seconds in total, corresponding to roughly $0.7$ milliseconds per segment. This suggests that MoTM remains computationally efficient at inference time, even on large-scale data. 
For comparison, SAITS requires full retraining on this out-of-distribution dataset and takes approximately 3h16 to reach the performance reported in \cref{tab:res-clean} on the same task.

\input{tables/table_inference_time}

%% file: tables/table_kernel_synth.tex
\begin{table}[b]
\centering
\caption{
Synthetic datasets generated by KernelSynth \cite{ansari2024chronos}.
SNR: Signal-to-Noise Ratio.
}
\setlength{\tabcolsep}{4pt}
\begin{tabular}{ccccccc}
\toprule
\multirow{2}{*}{\bfseries Dataset} & \multirow{2}{*}{\bfseries Samples} & \multirow{2}{*}{\bfseries Length} & \bfseries Sampling & \bfseries RBF & \multirow{2}{*}{\bfseries Period} & \bfseries Average \\
& & & \bfseries Freq. & \bfseries Scale & & \bfseries SNR (dB) \\
\midrule
\texttt{ks1D}   & 100 & 4032 & 1H    & 1.5  & 1D      & 20.6 \\
\texttt{ks1W}   & 100 & 5376 & 30min & 5    & 1W      & 22.3 \\
\texttt{ks1D1W} & 100 & 5376 & 15min & 1.25 & 1D + 1W & 14.9 \\
\bottomrule
\end{tabular}
\label{tab:kernel-synth}
\end{table}

%% file: tables/table_synthetic_res.tex
\begin{table}[h!]
\caption{Mean Absolute Errors (MAEs) on test series, z-normalized using the available context.
Best results are shown in \textbf{bold}, and second-best are \ul{underlined}.
MoTM results correspond to $\lambda = 2$.
MoTM improvement reports the relative improvement of MoTM over each baseline, averaged across all rows.
}
\setlength{\tabcolsep}{3pt}
\centering
\scalebox{0.77}{
\begin{tabular}{cc|ccc|cc|cc}
\toprule
& & \bfseries \multirow{2}{*}{\bfseries MoTM} & \multirow{2}{*}{\bfseries{Mixture I}} & \multirow{2}{*}{\bfseries {Mixture II}} & \bfseries {TimeFlow} & \bfseries {TimeFlow} & \bfseries \multirow{2}{*}{\bfseries {Linear}} & \multirow{2}{*}{\bfseries {Repeat}} \\
& & & & & \bfseries {1D} & \bfseries {1W} & &  \\
\midrule
\rowcolor{gray!15}
\multicolumn{9}{c}{\emph{MoTM In-Domain}} \\
\midrule
\multirow{4}{*}{\bfseries ks1D}
& \emph{Point 1} & 0.238 & 0.382 & \ul{0.232} & \bfseries 0.231 & 0.860 & 0.387 & 0.316 \\
& \emph{Point 2} & 0.246 & 0.389 & \ul{0.237} & \bfseries 0.236 & 0.868 & 0.530 & 0.317 \\
& \emph{Block 1} & 0.232 & 0.386 & \ul{0.228} & \bfseries 0.227 & 0.889 & 1.157 & 0.317 \\
& \emph{Block 2} & 0.232 & 0.389 & \bfseries{0.228} & \bfseries 0.228 & 0.894 & 1.154 & 0.316 \\
\midrule

\multirow{4}{*}{\bfseries ks1W}
& \emph{Point 1} & 0.122 & 0.309 & \bfseries 0.120 & 0.834 & \ul{0.121} & 0.146 & 0.292 \\
& \emph{Point 2} & 0.126 & 0.318 & \bfseries 0.123 & 0.849 & \bfseries 0.123 & 0.149 & 0.459 \\
& \emph{Block 1} & \bfseries 0.122 & 0.338 & \ul{0.124} & 0.891 & 0.127 & 0.217 & 0.164 \\
& \emph{Block 2} & \bfseries 0.123 & 0.349 & \ul{0.130} & 0.901 & 0.134 & 0.233 & 0.172 \\
\midrule
\multicolumn{2}{c|}{\bfseries MoTM improvement}
& \textbf{0.0\%} 
& \textbf{50.4\%} 
& \textcolor{purple}{\bfseries -1.0\%} 
& \textbf{44.4\%} 
& \textbf{40.4\%} 
& \textbf{48.7\%} 
& \textbf{36.0\%} \\
\midrule
\rowcolor{gray!15}
\multicolumn{9}{c}{\emph{MoTM Out-Of-Domain}} \\
\midrule
 
\multirow{4}{*}{\bfseries ks1D1W}
& \emph{Point 1} & \bfseries 0.145 & 0.409 & 0.183 & 0.583 & 0.583 & \ul{0.154} & 0.743 \\
& \emph{Point 2} & \bfseries 0.148 & 0.412 & 0.195 & 0.585 & 0.585 & \ul{0.164} & 0.806 \\
& \emph{Block 1} & \bfseries 0.153 & 0.440 & \ul{0.201} & 0.617 & 0.624 & 0.778 & 0.654\\
& \emph{Block 2} & \bfseries 0.155 & 0.446 & \ul{0.214} & 0.619 & 0.628 & 0.780 & 0.661\\

\midrule

\multicolumn{2}{c|}{\bfseries MoTM improvement}
& \textbf{0.0\%} 
& \textbf{64.8\%} 
& \textbf{24.2\%} 
& \textbf{75.0\%} 
& \textbf{75.2\%} 
& \textbf{48.8\%} 
& \textbf{78.9\%} \\
\bottomrule
\end{tabular}
}
\label{tab:res-synthetic}
\end{table}

%% file: figures/fig-synth-plots.tex
\begin{figure}[h!]
\centering
\subfigure[\emph{Point 2}]{
\includegraphics[width=0.99\textwidth]{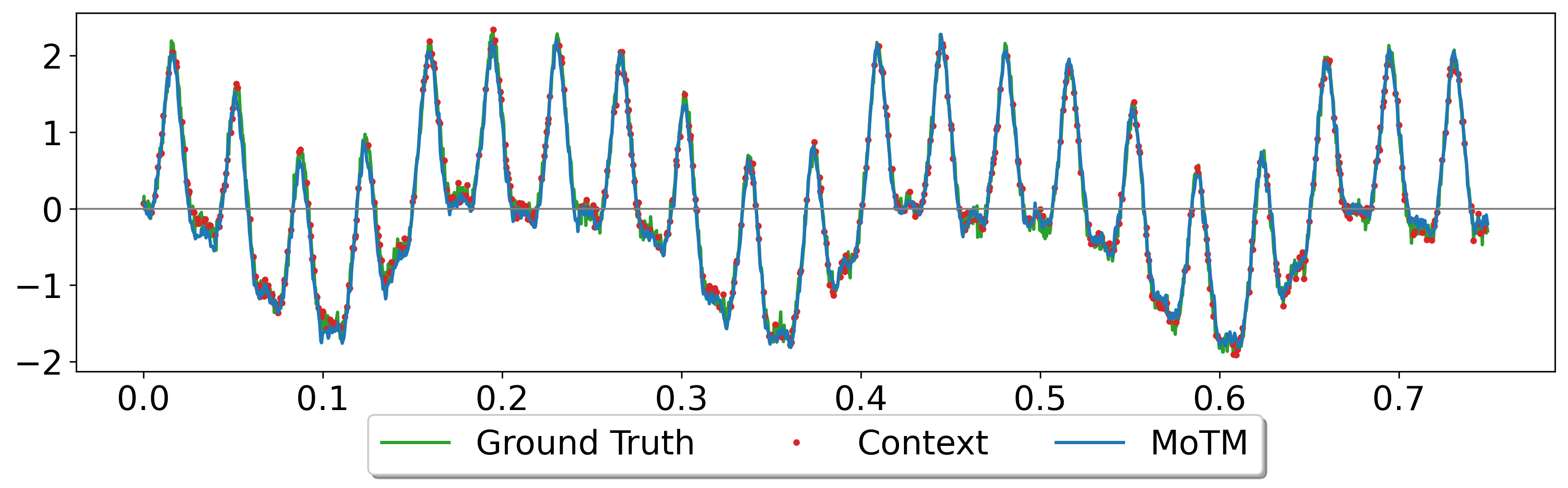}
}
\subfigure[\emph{Block 2}]{
\includegraphics[width=0.99\textwidth]{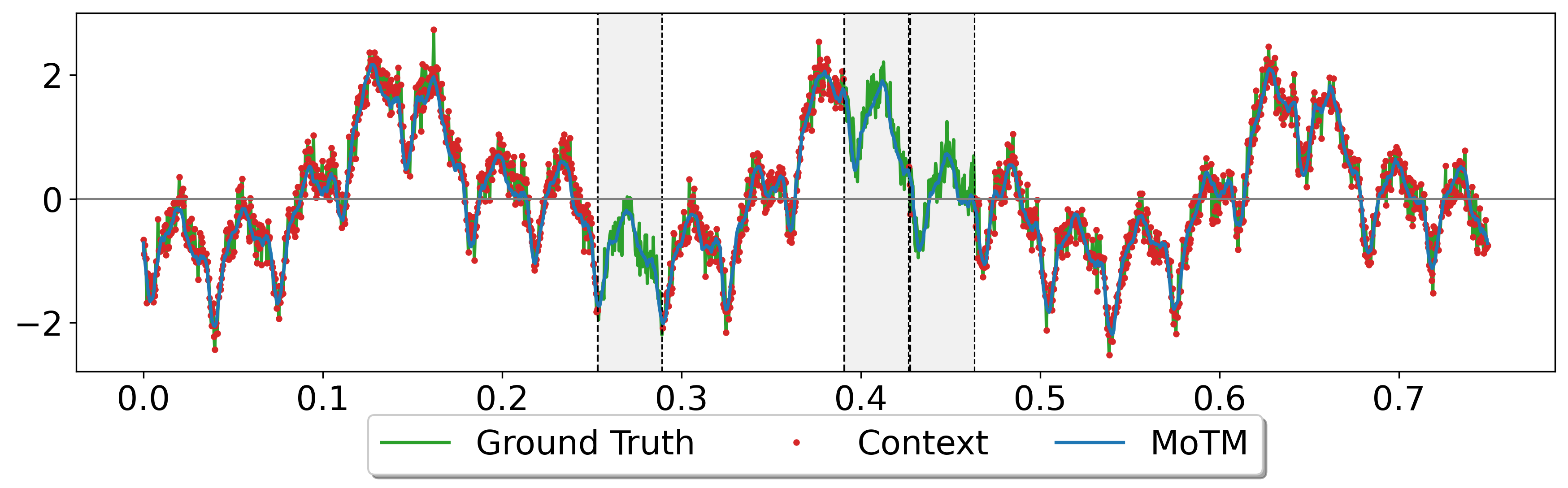}
}
\caption{
\emph{OOD \texttt{ks1D1W} dataset.}
MoTM performs imputation on (a) 70\% missing timesteps and (b) three one-day missing blocks.
Zoom on the first three weeks.
}
\label{fig:res-synthetic}
\end{figure}

%% file: tables/table_data.tex
\begin{table}[t]
\centering
\caption{
Datasets used for our experiments.
SNR: Signal-to-Noise Ratio.
}
\setlength{\tabcolsep}{4pt}
\resizebox{0.95\textwidth}{!}{
\begin{tabular}{l|cccc|cc}
\toprule
\multirow{2}{*}{\bfseries Dataset} & \multirow{2}{*}{\bfseries Samples} & \bfseries Total Series & \bfseries Sampling & \bfseries Average & \bfseries MoTM & \bfseries Input \\
 & & \bfseries Length &\bfseries Freq. & \bfseries SNR (dB) & \bfseries Splits & \bfseries Length\\
\midrule
\href{https://zenodo.org/records/4656140}{\texttt{Electricity}}
& 321 & 26 304 & 1H    & 24.6 & Train, Infer. & 672   \\ 
\href{https://zenodo.org/records/3889974}{\texttt{Solar}}
& 137 & 52 560 & 10min & 17.4 & Train, Infer. & 4 032 \\
\href{https://www.kaggle.com/datasets/nicholasjhana/energy-consumption-generation-prices-and-weather/data}{\texttt{SpanishW-T}}
& 5   & 35 000 & 1H    & 35.3 & Train, Infer. & 672   \\
\midrule
\href{https://zenodo.org/records/4656132}{\texttt{Traffic}}
& 861 & 17 544 & 1H    & 18.3 & Inference      & 672  \\
\href{https://github.com/zhouhaoyi/ETDataset}{\texttt{ETTh1}}
& 7   & 17 420 & 1H    & 11.8 & Inference      & 672  \\
\href{https://github.com/zhouhaoyi/ETDataset}{\texttt{ETTh2}}
& 7   & 17 420 & 1H    & 11.6 & Inference      & 672  \\
\href{https://drive.google.com/drive/folders/1ohGYWWohJlOlb2gsGTeEq3Wii2egnEPR}{\texttt{Weather}}
& 11  & 35 064 & 1H    & 14.7 & Inference      & 672  \\
\href{https://www.kaggle.com/datasets/nicholasjhana/energy-consumption-generation-prices-and-weather/data}{\texttt{Spanish E}}
& 9   & 35 064 & 1H    & 22.0 & Inference      & 672  \\
\bottomrule
\end{tabular}
}
\label{tab:data}
\end{table}

%% file: tables/table_v3.tex
\begin{table}[]
\centering
\setlength{\tabcolsep}{3pt}
\caption{
MAEs on the test fraction of the real world datasets.
Best performance emphasized in \textbf{bold}, second best \ul{underlined}.
MoTM improvement reports the relative improvement of MoTM over each baseline, averaged across all rows.
}
\resizebox{0.95\textwidth}{!}{
\begin{tabular}{cc|cc|ccc|cc}
\toprule
&& \multicolumn{2}{c|}{\emph{Zero-shot}} & \multicolumn{3}{c|}{\emph{Supervised}} & \multicolumn{2}{c}{\emph{Statistical}} \\
\midrule
& & \bfseries MoTM & \bfseries MOMENT & \bfseries TimeFlow & \bfseries BRITS & \bfseries SAITS & \bfseries Linear & \bfseries Repeat 
\\
\midrule
\rowcolor{gray!15}
\multicolumn{9}{c}{\emph{MoTM In-Domain}} \\
\midrule
\multirow{4}{*}{\bfseries Electricity} 
& \emph{Point 1} & \bfseries 0.196 & 0.861 & 0.274 & 0.324 & \ul{0.211} & 0.306 & 0.334 \\
& \emph{Point 2} & \bfseries 0.229 & 0.863 & 0.322 & 0.465 & \ul{0.258} & 0.435 & 0.357 \\
& \emph{Block 1} & \bfseries 0.257 & 0.478 & \ul{0.291} & 0.522 & 0.296 & 1.025 & 0.312 \\
& \emph{Block 2} & \bfseries 0.259 & 0.538 & 0.298 & 0.465 & \ul{0.292} & 1.027 & 0.313 \\
\midrule

\multirow{4}{*}{\bfseries Solar} 
& \emph{Point 1} & 0.083 & 0.857 & 0.085 & \ul{0.072} & 0.077 & \bfseries 0.036 & 0.265 \\
& \emph{Point 2} & 0.092 & 0.858 & 0.097 & \ul{0.083} & 0.130 & \bfseries 0.055 & 0.281 \\
& \emph{Block 1} & \ul{0.253} & 0.755 & 0.257 & 0.308 & 0.361 & 0.883 & \bfseries 0.244 \\
& \emph{Block 2} & \ul{0.256} & 0.781 & 0.258 & 0.314 & 0.355 & 0.889 & \bfseries 0.244 \\
\midrule

\multirow{4}{*}{\bfseries Spanish W-T} 
& \emph{Point 1} & 0.214 & 0.835 & 0.283 & 0.373 & \ul{0.205} & \bfseries 0.169 & 0.520 \\
& \emph{Point 2} & \bfseries 0.253 & 0.838 & 0.309 & 0.473 & 0.295 & \ul{0.277} & 0.585 \\
& \emph{Block 1} & \ul{0.402} & 0.511 & \bfseries 0.391 & 0.685 & 0.444 & 0.889 & 0.484 \\
& \emph{Block 2} & \ul{0.404} & 0.548 & \bfseries 0.396 & 0.689 & 0.451 & 0.898 & 0.470 \\
\midrule
\multicolumn{2}{c|}{\bfseries MoTM improvement}
& \textbf{0.0\%} 
& \textbf{62.8\%} 
& \textbf{10.8\%} 
& \textbf{30.6\%} 
& \textbf{12.6\%} 
& \textbf{22.5\%} 
& \textbf{32.2\%} \\
\midrule
\rowcolor{gray!15}
\multicolumn{9}{c}{\emph{MoTM Out-of-Domain}} \\
\midrule

\multirow{4}{*}{\bfseries Traffic} 
& \emph{Point 1} & 0.246 & 0.770 & \ul{0.240} & 0.267 & \bfseries 0.201 & 0.287 & 0.379 \\
& \emph{Point 2} & 0.294 & 0.774 & \ul{0.291} & 0.374 & \bfseries 0.241 & 0.421 & 0.416 \\
& \emph{Block 1} & \ul{0.313} & 0.478 & 0.389 & 0.415 & \bfseries 0.227 & 0.983 & 0.340 \\
& \emph{Block 2} & \ul{0.318} & 0.521 & 0.395 & 0.431 & \bfseries 0.231 & 0.985 & 0.341 \\
\midrule

\multirow{4}{*}{\bfseries ETTh1} 
& \emph{Point 1} & \ul{0.340} & 0.812 & 0.410 & 0.539 & 0.347 & \bfseries 0.334 & 0.594 \\
& \emph{Point 2} & \bfseries 0.389 & 0.814 & 0.482 & 0.633 & \ul{0.421} & 0.426 & 0.635 \\
& \emph{Block 1} & \bfseries 0.490 & 0.633 & 0.553 & 0.723 & \ul{0.535} & 0.845 & 0.558 \\
& \emph{Block 2} & \bfseries 0.488 & 0.664 & 0.557 & 0.730 & \ul{0.536} & 0.834 & 0.559 \\
\midrule

\multirow{4}{*}{\bfseries ETTh2} 
& \emph{Point 1} & 0.442 & 0.806 & 0.489 & 0.533 & \ul{0.422} & \bfseries 0.406 & 0.763 \\
& \emph{Point 2} & 0.496 & 0.806 & 0.521 & 0.610 & \ul{0.486} & \bfseries 0.471 & 0.805 \\
& \emph{Block 1} & \ul{0.609} & 0.704 & 0.631 & 0.691 & \bfseries 0.602 & 0.761 & 0.738 \\
& \emph{Block 2} & \bfseries 0.600 & 0.704 & 0.619 & 0.722 & \ul{0.610} & 0.760 & 0.716 \\
\midrule

\multirow{4}{*}{\bfseries Weather} 
& \emph{Point 1} & 0.326 & 0.816 & 0.330 & 0.389 & \ul{0.287} & \bfseries 0.260 & 0.739 \\
& \emph{Point 2} & 0.375 & 0.819 & 0.383 & 0.484 & \ul{0.351} & \bfseries 0.323 & 0.803 \\
& \emph{Block 1} & \bfseries 0.524 & 0.640 & 0.627 & 0.689 & \ul{0.584} & 0.621 & 0.677 \\
& \emph{Block 2} & \bfseries 0.527 & 0.669 & 0.633 & 0.691 & \ul{0.582} & 0.620 & 0.682 \\
\midrule

\multirow{4}{*}{\bfseries Spanish E} 
& \emph{Point 1} & 0.235 & 0.818 & 0.329 & 0.311 & \ul{0.189} & \bfseries 0.164 & 0.678 \\
& \emph{Point 2} & 0.286 & 0.823 & 0.376 & 0.425 & \ul{0.285} & \bfseries 0.253 & 0.738 \\
& \emph{Block 1} & \ul{0.507} & 0.622 & \bfseries 0.503 & 0.675 & 0.536 & 0.606 & 0.644 \\
& \emph{Block 2} & \ul{0.503} & 0.649 & \bfseries  0.499 & 0.675 & 0.535 & 0.604 & 0.633 \\
\midrule

\multicolumn{2}{c|}{\bfseries MoTM improvement}
& \textbf{0.0\%} 
& \textbf{40.3\%} 
& \textbf{10.2\%} 
& \textbf{24.0\%} 
& \textcolor{purple}{\bfseries -5.7\%} 
& \textbf{13.2\%} 
& \textbf{31.1\%} \\
\bottomrule

\end{tabular}
}
\label{tab:res-clean}
\end{table}

%% file: plots/fig-energy.tex
\begin{figure}
    \centering
    \includegraphics[width=0.99\linewidth]{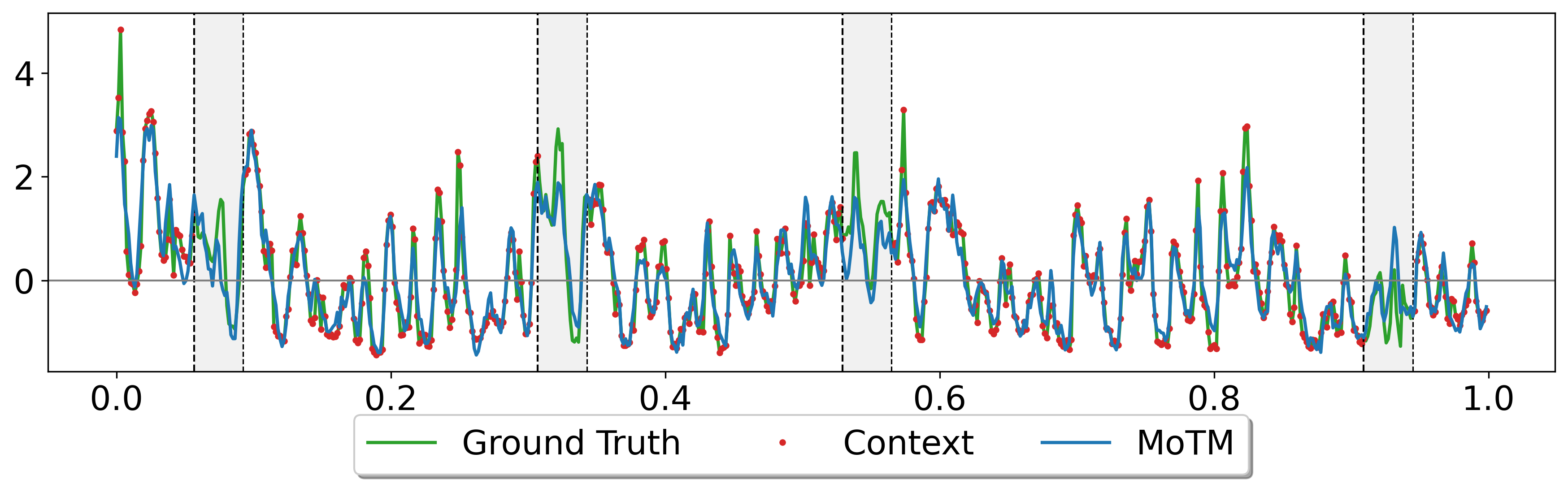}
    \caption{\emph{OOD \texttt{SpanishE} dataset.} MoTM imputation on four one-day missing blocks.}
    \label{fig:spanishE-vizu}
\end{figure}

%% file: figures/fig-ablation_ntrain.tex
\begin{figure}[h!]
\centering
\subfigure[Electricity]{\includegraphics[width=0.225\textwidth]{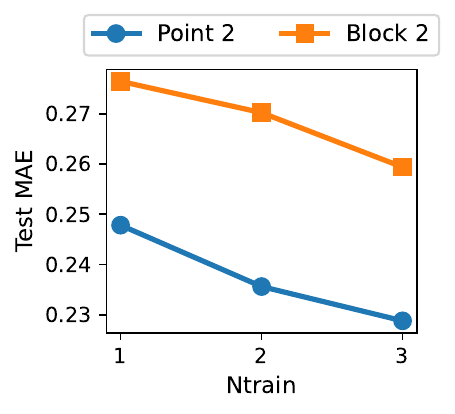}}
\subfigure[Traffic]{\includegraphics[width=0.225\textwidth]{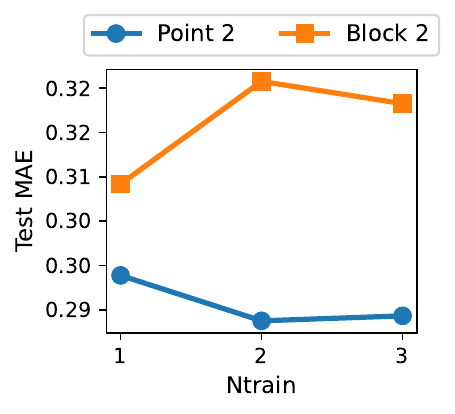}}
\subfigure[ETTh1]{\includegraphics[width=0.225\textwidth]{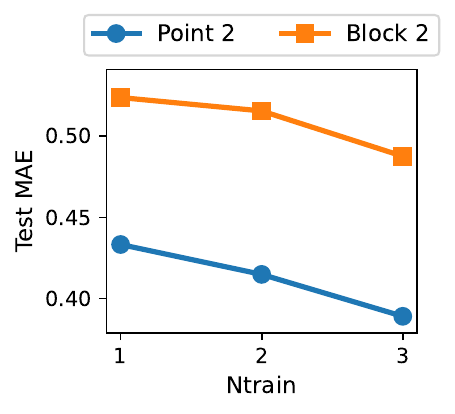}}
\subfigure[Weather]{\includegraphics[width=0.225\textwidth]{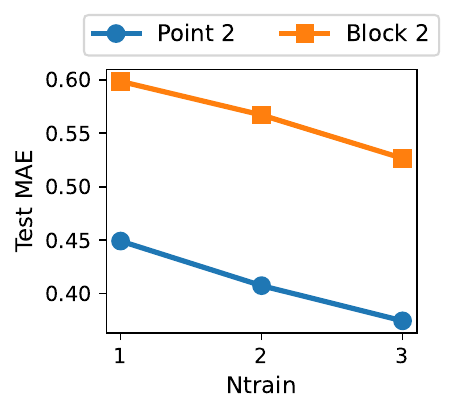}}
\caption{
\emph{Ablation: number of basis components.}
Test MAE scores on one ID (\texttt{Electricity}) and three OOD (\texttt{Traffic}, \texttt{ETTh1}, \texttt{Weather}) datasets for a mixture of $N_{\text{train}}\in\{1,\,2,\,3\}$ TimeFlow components and ridge orchestrator.
The mixture components are obtained by successively training on \texttt{Electricity}, \texttt{Solar} and \texttt{SpanishW-T}.
}
\label{fig:ablation}
\end{figure}

%% file: tables/table_inference_time.tex
\begin{table}[]
\caption{Computation time on the \texttt{Traffic} test dataset (NVIDIA H100 GPU). MoTM is evaluated in a zero-shot setting, while SAITS requires training.}
\setlength{\tabcolsep}{4pt} 
\centering
\begin{tabular}{lccc}
\toprule 
\bfseries Method & \bfseries Segments & \bfseries Sequence length & \bfseries Compute time \\
\midrule
MoTM (zero-shot) & 83,517 & 672 & 61s \\
SAITS (training + inference) & 83,517 & 672 & 3h16 \\
\bottomrule
\end{tabular}
\label{tab:infer-time}
\end{table}

%% file: 05-conclusion.tex
\section{Conclusion and Discussion}

In this work, we introduced MoTM, a mixture-based architecture that extends the capabilities of continuous-time TimeFlow models to a zero-shot imputation setting. By aggregating specialized TimeFlow models trained on distinct distributions, MoTM effectively handles a wide range of missing data patterns and sampling rates, without the need for retraining. Our experiments confirm its strong generalization performance across both synthetic and real-world datasets, especially in out-of-distribution scenarios.

While MoTM inference is slightly slower than single-model approaches due to its mixture structure, it remains efficient relative to the substantial training time required by comparable models retrained from scratch, as shown in \cref{tab:infer-time}. Overall, the performance gains brought by MoTM are not uniform across all ID datasets. For instance, the \texttt{Solar} dataset shows limited improvement, highlighting that the benefit of model mixing depends on the underlying data distribution. Identifying the most effective combination of models remains thus an open challenge. Moreover, the supervised baseline \texttt{SAITS} outperforms MoTM in several settings, suggesting that there is still room for improvement. Future work will explore the construction of large, unified databases enabling the training of TimeFlow models at scale for diverse distributions, and the integration of more expressive orchestration mechanisms to further enhance inference quality.

\section{Reproducibility}

All code and data used in this work are publicly available to ensure reproducibility of the results.  
\begin{itemize}
    \item Code repository: \url{https://github.com/EtienneLnr/MoTM}
    \item Datasets: \url{https://zenodo.org/records/17177008}
\end{itemize}

\section{Acknowledgements}

We would like to thank Louis Serrano for his insightful discussions about this paper. We would also like to thank Adrien Petralia and Camille Georget for their helpful feedback and proofreading assistance.